\documentclass[10pt,twocolumn,letterpaper]{article}

\usepackage{wacv}
\usepackage{times}
\usepackage{epsfig}
\usepackage{graphicx}
\usepackage{amsmath}
\usepackage{amssymb}
\usepackage{array}
\usepackage{subfig}
\usepackage{multirow}
\usepackage[normalem]{ulem}
\usepackage{float}
\usepackage{textcomp}
\usepackage{svg}
\usepackage{authblk}
\newcolumntype{P}[1]{>{\centering\arraybackslash}p{#1}}
\newcommand{\rowSpace}{\vspace{-0.4cm}}
\usepackage[pagebackref=true,breaklinks=true,letterpaper=true,colorlinks,bookmarks=false]{hyperref}
\wacvfinalcopy % *** Uncomment this line for the final submission

\graphicspath{ {images/} }

\def\wacvPaperID{104} % *** Enter the wacv Paper ID here

\ifwacvfinal\fi
\begin{document}

%%%%%%%%% TITLE
\title{EpO-Net: Exploiting Geometric Constraints on Dense Trajectories for Motion Saliency}
% Authors at the same institution
%\author{First Author \hspace{2cm} Second Author \\
% Authors at different institutions
% \author{First Author \\
% \and
% Second Author \\
\author[1]{Muhammad Faisal}
\author[2]{Ijaz Akhter}
\author[1]{Mohsen Ali}
\author[3]{Richard Hartley}
\affil[1]{Information Technology University, Pakistan} 
\makeatletter
\renewcommand\AB@affilsepx{, \protect\Affilfont}
\makeatother
\affil[2]{KeepTruckin, Inc} \affil[3]{Australian National University,  Australia
% \authorcr {\tt\{m.faisal, mohsen.ali\}@itu.edu.pk, ijaz.akhter@keeptruckin.com, richard.hartley@anu.edu.au}
}

\maketitle
\ifwacvfinal\thispagestyle{empty}\fi

%%%%%%%%% ABSTRACT

% Repharse: features? correspondances
\begin{abstract}
   The existing approaches for salient motion segmentation are unable to explicitly learn geometric cues and often give false detections on prominent static objects. We exploit multiview geometric constraints to avoid such shortcomings. To handle the nonrigid background like a sea, we also propose a robust fusion mechanism between motion and appearance-based features. We find dense trajectories, covering every pixel in the video, and propose trajectory-based epipolar distances to distinguish between background and foreground regions. Trajectory epipolar distances are data-independent and can be readily computed given a few features' correspondences between the images. We show that by combining epipolar distances with optical flow, a powerful motion network can be learned. Enabling the network to leverage both of these features, we propose a simple mechanism, we call input-dropout. Comparing the motion-only networks, we outperform the previous state of the art on DAVIS-2016 dataset by 5.2\% in the mean IoU score. By robustly fusing our motion network with an appearance network using the input-dropout mechanism, we also outperform the previous methods on DAVIS-2016, 2017 and Segtrackv2 dataset.
\end{abstract}

\rowSpace

\section{Introduction}
\setlength{\tabcolsep}{0.25pt}
\begin{figure}[t]    
% \begin{table}
\center
\begin{tabular}{ccc}
\subfloat{\includegraphics[width=2.75cm,height=1.7cm]{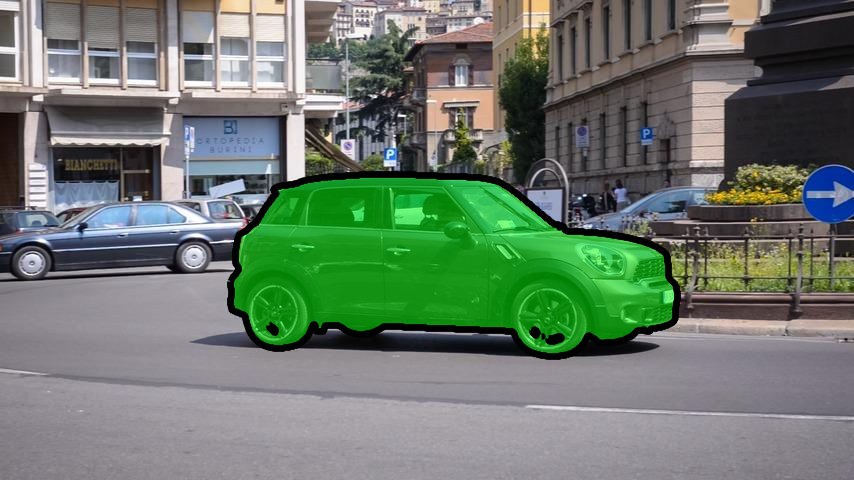}} & \subfloat{ \includegraphics[width=2.75cm,height=1.7cm]{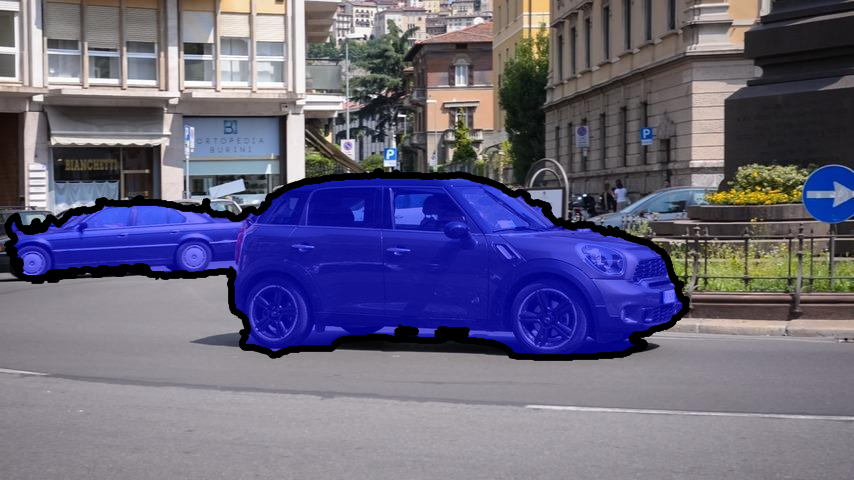}} & \subfloat{ \includegraphics[width=2.75cm,height=1.7cm]{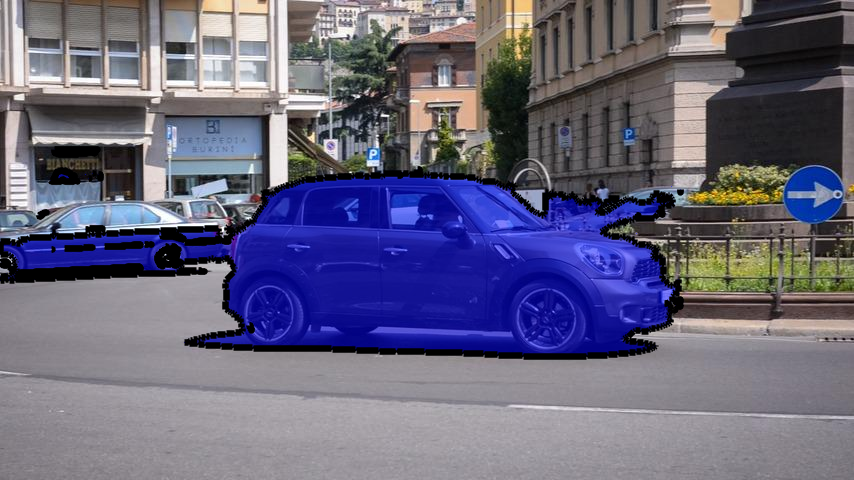}} \\
Ground truth & LVO~\cite{visMem} & STP~\cite{STP} \\
\subfloat{ \includegraphics[width=2.75cm,height=1.7cm]{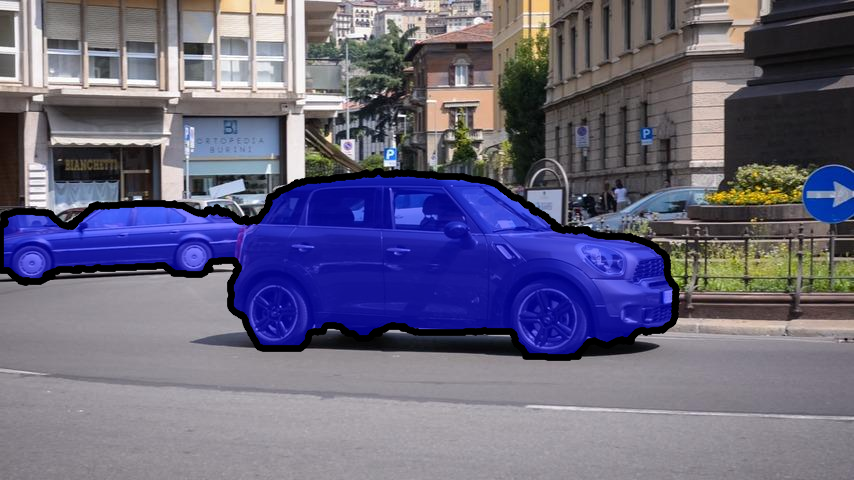}} & \subfloat{ \includegraphics[width=2.75cm,height=1.7cm]{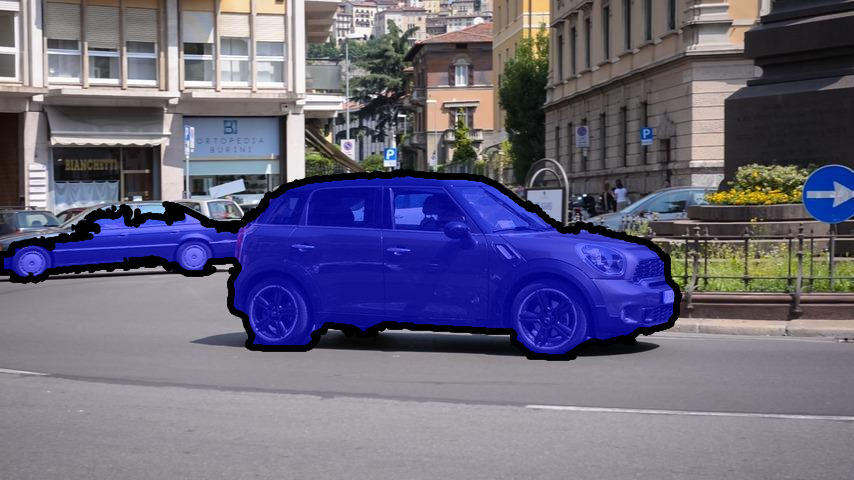}} & \subfloat{ \includegraphics[width=2.75cm,height=1.7cm]{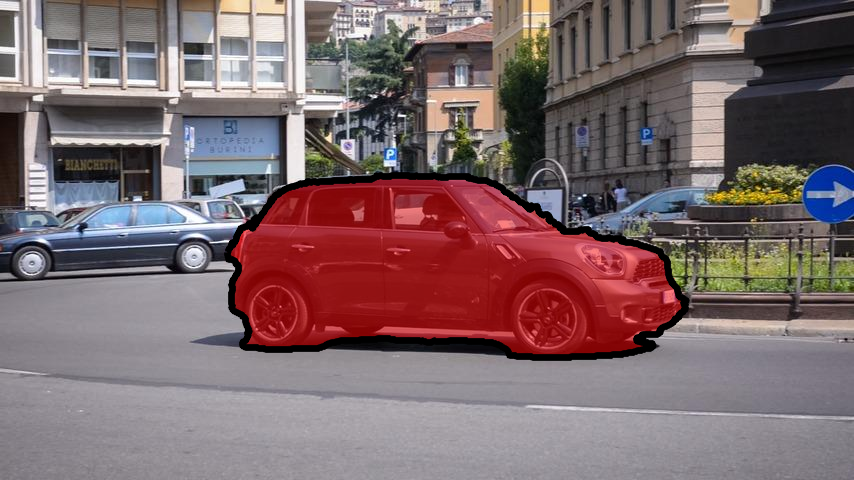}} \\
MotAdapt~\cite{MotAdapt} & AGS~\cite{ags} & Our \\
\end{tabular}
\caption{Existing methods fail to automatically learn geometric cues between the foreground objects and the rigid background. As a result, they often give false detections on prominent static objects, as shown here in an example from DAVIS \cite{DAVIS2016}. Whereas by exploiting these constraints over the whole video, we avoid making such mistakes.}
\label{fig:teaser}
\rowSpace
% \end{table}
\end{figure}

Segmenting object(s) with significant motion in a video is called Salient Motion Segmentation. In contrast, segmenting the most prominent object(s) in an image (or a video) is Salient Appearance Segmentation. While the data-driven approaches have been quite successful for the later, we argue, that the former suffers from the scarcity of the video-based training data and remains ill-posed. 
Specifically, for a moving camera, it remains hard to learn, whether the 2D projected motion field corresponds to a static object in the video, or the one having independent motion.
To segment out the rigid background from the independently moving foreground objects, we exploit extensively studied geometric constraints \cite{hartley2003multiple}, over the complete video, in a learning paradigm. Unlike the data-dependent learning, these constraints have closed-form solutions and can be computed very efficiently. 
% Our method can still handle nonrigid background with the fusion of motion and appearance-based features. 
In Fig.~\ref{fig:teaser} we give an example from DAVIS \cite{DAVIS2016}, showing that the previous approaches give false detections on prominent static objects; whereas the proposed approach can disambiguate static and nonstatic objects. This clearly shows that the existing deep-networks are unable to automatically learn the geometric cues even when the optical flow was provided as an input.

To exploit multiview geometric constraints, we convert optical flow between consecutive frames into dense trajectories, covering every pixel in the video, and then use trifocal tensors to find epipolar distances \cite{hartley2003multiple} for them. The trajectory epipolar distance serves as a measure of (non)rigidity: a small distance corresponds to the rigid background, and a large distance implies a foreground object(s). 

Trajectory epipolar distances,  capture temporally global constraint on the foreground and background region, whereas optical flow only captures local temporal information.
% However, the former is quite sensitive to optical flow errors. 
In essence, they both are complementary and by combining them, powerful features for motion saliency can be learned. Given trajectory epipolar distances and optical flow as an input, we build an encoder-decoder based network \cite{uNet}, called EpO-Net. We devise a strategy called input-dropout, enabling the network to learn robust motion features and handle failure cases of one of the two inputs.

EpO-Net brings two key advantages over the existing motion network, Mp-Net \cite{MpNet}. 1) EpO-Net exploits geometric constraints over a large temporal window, whereas Mp-Net makes suboptimal decisions based on temporally local optical flow information. Consequently, as we show, EpO-Net can be trained on smaller training data, while having better generalization than Mp-Net. 2) In contrast to Mp-Net, EpO-Net does not require any objectness score on top of the estimated motion saliency map. The main reason for this is, we prepare and train our network on a more realistic but synthetic data consisting of real backgrounds and synthetic foreground objects, called Real Background and Synthetic Foreground (RBSF) dataset. Whereas, Mp-Net was trained on unrealistic synthetic 3D flying objects \cite{F3DT}.

Being a motion-only network, EpO-Net cannot handle a nonrigid background. To handle this case, we exploit appearance \cite{deeplab} along with motion-based features in the form of a joint network, EpO-Net+.
Using the proposed input-dropout strategy, we show that the EpO-Net+ is robust against the failure cases of individual motion and appearance-based features.

To the best of our knowledge, ours is the first method to combine geometric constraints in a learning paradigm for motion segmentation. Our paper has three main contributions. 1) A motion only network based on trajectory epipolar distance and optical flow. 2) Our RBSF dataset, that can be used to train salient motion segmentation. Applications like video annotation \cite{feng2004multiple}, object tracking \cite{yilmaz2006object}, and video anomaly detection \cite{xiang2008video}, can use our network and the dataset, to exploit geometric constraints on the rigid world. The source code of our method, as well as the dataset, is publicly released\footnote{\url{https://github.com/mfaisal59/EpONet}}.
%We would release the source code of our method as well as the dataset to the general public. 
3) The input-dropout technique, which can be used to robustify early or late fusion of features in deep architectures. Our motion network outperforms Mp-Net on DAVIS-2016 \cite{DAVIS2016} by a significant margin of 5.2\% in mean IoU score and is quite close to other recent methods exploiting additional appearance features. The proposed joint network also demonstrates significant improvement over the previous methods on DAVIS (2016 \cite{DAVIS2016} \& 2017 \cite{DAVIS2017}) and Segtrack-v2 \cite{segtrackv2}.

\section{Related Work}
Recently, video object segmentation (VOS) has been gaining interest \cite{fusionseg,MpNet, visMem,PDB,STP,SegFlow,ARP}, much credit to the new challenging benchmark datasets. 
One of the factors to categorize existing approaches could be the degree of supervision.
Supervised approaches \cite{NagarajaSB15, ChenPMG18} or interactive ones assume user input, in the form of scribbles, is available at multiple instances, helping algorithm refine the results. 
Semi-Supervised methods \cite{STP,MoNetECCV2018, videoMatchECCV2018, instanceEmbTrans2018, CNN_MRF, osvosJournal, PReMVOS}, assume that at least for the first frame, segmentation is given, reducing the problem to label propagation. 
For brevity, we discuss only a few prominent unsupervised methods.

In unsupervised settings, to make the problem tractable the motion-saliency constraint is enforced. 
Many methods try to capture motion information across the multiple frames, mostly by constructing long sparse point trajectories  ~\cite{brox2010object, fragkiadaki2012video, ochs2012higher, shi1998motion}. Salient object segmentation is then reduced to clustering these trajectories \cite{CUT} and converting them into dense points~\cite{MSG}.
Among the other early methods, few methods ~\cite{KEY, segtrackv2, ma2012maximum, zhang2013video, perazzi2015fully} extract object proposals ~\cite{endres2010category} and try to build the connection between the proposals temporally. 
These trajectory based methods are not robust because they heavily rely on feature matching, that may fail due to occlusion, fast motion, and appearance change.

Recently deep learning based methods have been used to solve the VOS problem. 
Broadly, these techniques have three components: 1) network to capture the motion, 2) extract appearance information, 3) a temporal memory so that the decision made at one frame is propagated to the others~\cite{visMem, fusionseg, SegFlow, PDB}.
Among all these approaches, Mp-Net \cite{MpNet} and LVO \cite{visMem} are very close to our method. Mp-Net constructs an encoder/decoder based network to segment the optical flow into the salient and non-salient one. 
Encoder/decoder network is trained on large synthetic dataset \cite{F3DT} and then fine-tuned on DAVIS \cite{DAVIS2016}.
Since motion information they learn is not sufficient, they rely on an objectness score \cite{objectness} to clean their results. 
LVO, builds on Mp-Net, using bi-directional ConvGRU to propagate the information across the other frames. Their results improve drastically (LSMO \cite{lsmo}) by just using a better optical flow estimation and appearance model (DeepLabv2 instead of Deep Lab v1). 
MotAdapt \cite{MotAdapt} used the teacher-student learning paradigm, where the teacher provides pseudo labels using the optical flow and the image as input.  

AGS \cite{ags} explores the concepts of video saliency or dynamic fixation prediction, with an argument that unsupervised VOS is closely related to the video saliency \cite{wang2018revisiting}. Authors trained a visual attention module on the dynamic fixation data, collected by tracking viewers' eyes while they watch videos. 
Unlike AGS which required the data gathered by tracking the viewer's gaze, we try to model the concept of motion-saliency by exploiting the geometric constraints inside the video itself and do not require extra data.  

An early method by Torr \cite{torr1998geometric}, Sheikh et. al. \cite{sheikh2009background}, and Tron and Vidal \cite{tron2007benchmark}, try to exploit motion models. \cite{sheikh2009background}, and \cite{tron2007benchmark} exploited trajectory information to separate out the foreground and background objects. 
Many recent methods \cite{ARP,cliqu2018, STP} have relied on the previous trajectory-based segmentation work, using the deep features for image saliency and optical flow for motion saliency to construct a neighborhood graph. \cite{shankar2015video} used optical flow-based point trajectories to propagate the user input scribbles. \cite{wang2017super} clustered neighboring trajectories to create super-trajectories, and tracked the mask, provided as input, in the first frame of the video. However, they have not exploited the geometry-based constraints, rather rely on the heuristics and complex pipeline.

Our work relies on all the three techniques. We use optical flow to build trajectories and geometry-based technique to penalize the trajectories not following the geometric constraint.
To make our deep learning models robust, we design the input-dropout technique for the training.
To the best of our knowledge, we are the first one to combine CNNs and geometrical constraints for VOS. 

\section{Epipolar Constraints on Dense Trajectories}
\label{epipolarDistanceSection}

\begin{figure}[t]
\centering
\includegraphics[width=\columnwidth]{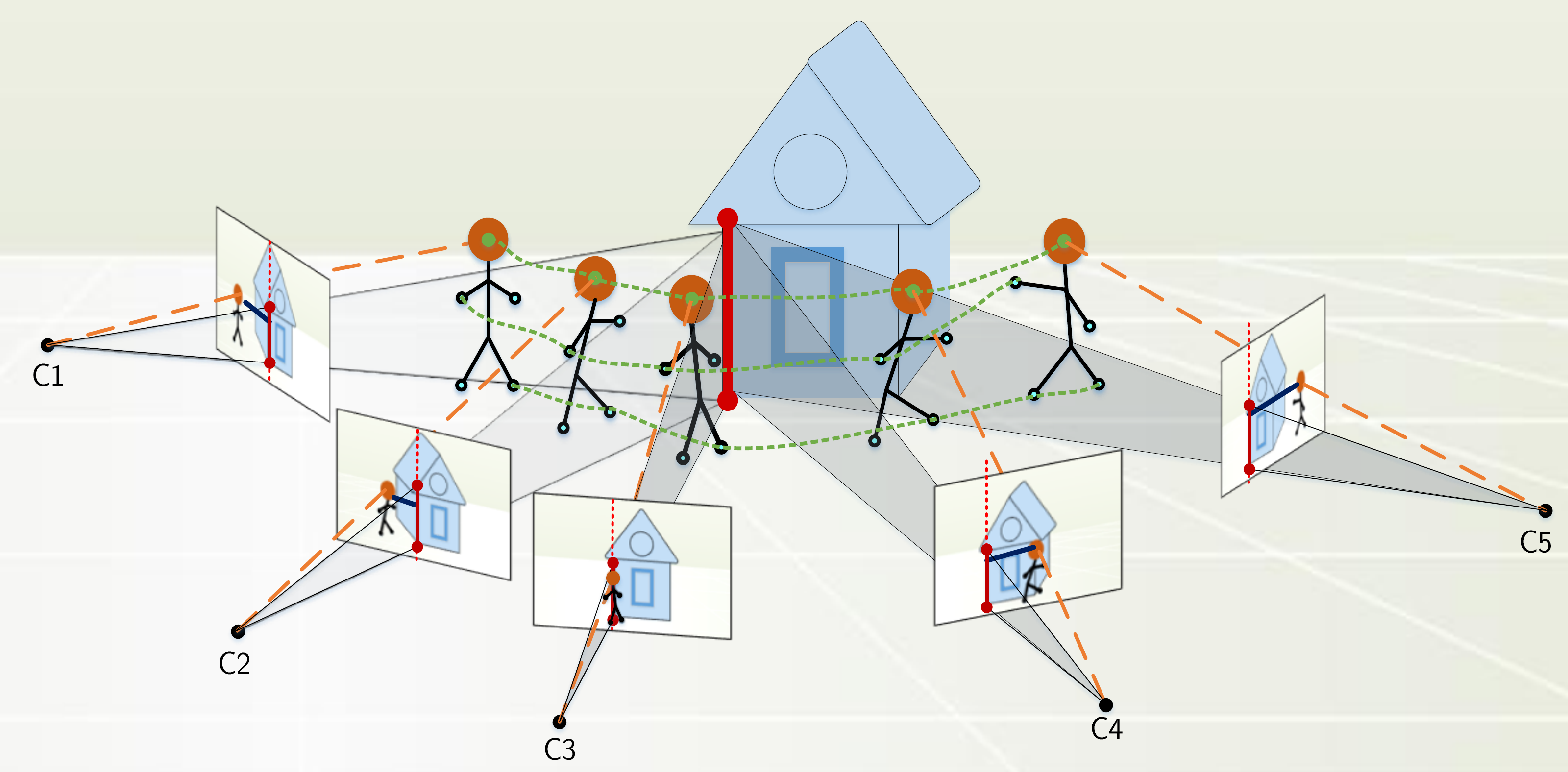}
\caption{An illustration of multiview geometric constraints on rigid points. A 3D rigid line (red) is viewed by a moving camera. The projections of its 2D projections in 3D should meet at the actual line. In contrast, the 2D projections of a 3D nonrigid point (orange) are not constrained to lie on any 3D lines. This relationship can be captured in the form of trifocal tensors (or fundamental matrices) in the frames. In contrast to rigid points, the nonrigid point may not lie on the corresponding epipolar lines and their epipolar distances can serve as a measure of nonrigidity.}
\label{fig:geometryFig}
\vspace{-0.5cm}
\end{figure}

Existing methods for salient motion segmentation, use appearance, and optical flow based features to distinguish foreground from background. These features are not geometry inspired, learned from the data and alone do not provide enough constraints for the rigid background. We propose geometry inspired features and leverage them in a learning pipeline. We use trifocal tensors to constraint the rigid background in the video and propose epipolar distances for the dense trajectories as a measure of nonrigidity (See Fig. \ref{fig:geometryFig}).

We first find forward and backward optical flow of $F$ frames, each of height $h$ and width $w$, using \cite{chen2016full} and then convert it into $T$ dense trajectories covering every pixel in the video. Each trajectory, $\mathbf{X}^i$, where $i \in \{1,\hdots,T\}$, is an $F\times1$ vector of 2D image coordinates and may consists of missing values due to pixels' occlusion. $T~\gg~hw$, because for every occlusion new pixels appear. 
We use forward and backward optical flow consistency to find occluding regions.
We stack all the trajectories into a $F~\times~T$ sparse matrix, $\mathbf{X}$. 

\begin{figure}[t]
\centering
\includegraphics[width=\columnwidth]{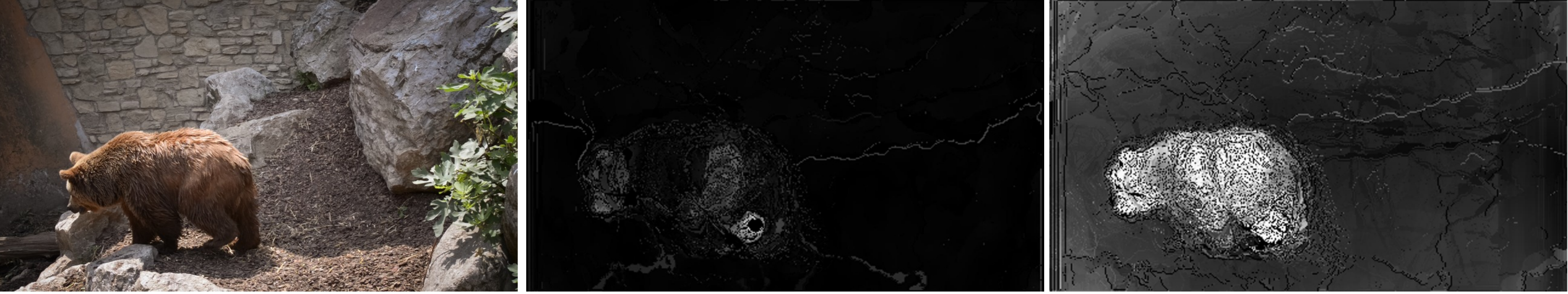}
\caption{An illustration of exploiting the complete trajectories to find epipolar distances. Part of the bear remains static in this and the previous frame, giving small epipolar distance (middle). Since trajectories aggregate these distances over their full time-span, the trajectory-based epipolar distances are still high for almost the complete bear (right).}
\vspace{-0.5cm}
\label{fig:epipolarDistance}
\end{figure}

Once trajectories are found, we estimate the dominant rigid background, by finding the trifocal tensors in every three consecutive frames, using the six-point algorithm \cite{hartley2003multiple}\footnote{Algorithm 20.1 page 511, Hartley \& Zisserman (2nd Ed)}, and RANSAC. We convert the trifocal tensor to the corresponding six pair-wise fundamental matrices, $\mathbf{F}_{12}, \mathbf{F}_{21}, \mathbf{F}_{13}, \mathbf{F}_{31}, \mathbf{F}_{23}, \mathbf{F}_{32}$ \cite{hartley2003multiple}\footnote{Algorithm 15.1, page 375, Hartley \& Zisserman (2nd Ed)}. When the camera is static and optical flow is zero for the background, the estimation of the trifocal tensor can become degenerate. Any skew-symmetric matrix, in this case, would be a valid fundamental matrix. To avoid degeneracy, we first detect if the camera remains static, by checking if at least 50\% of the pixels have zero optical flow, in the current triplet of frames. Then we initialize fundamental matrices to arbitrary skew-symmetric matrices. 

We find the epipolar distances for the triplet as follows. Let $\mathbf{x}_{j1}, \mathbf{x}_{j2}$ and $\mathbf{x}_{j3}$ denote the homogenous 2D coordinates of the selected three frames in the $j^\mathrm{th}$ trajectory. We find the distance between $\mathbf{x}_{j1}$ and $\mathbf{x}_{j2}$ as,
\vspace{-0.1cm}
\begin{align}
    \mathbf{l}_{21}  & =  \mathbf{F}_{21} \mathbf{x}_{j1},\\
      d_{j12}& = \mathbf{x}_{j2}^T  \mathbf{l}_{21}/ \sqrt{\mathbf{l}_{21}(1)^2 + \mathbf{l}_{21}(2)^2},
\end{align}
where $\mathbf{l}_{21}$ is the epipolar line in frame 2 corresponding to the frame 1, $\mathbf{l}_{21}(i)$, its $i^\mathrm{th}$ component and $d_{j12}$ is the distance between the line and $\mathbf{x}_{j2}$. By normalizing the line w.r.t its magnitude, gives the normlize epipolar distance. The triplet epipolar distance would be
\vspace{-0.1cm}
\begin{equation}
d_{j123} = d_{j12}+d_{j21}+d_{j13}+d_{j31}+d_{j23}+d_{j32}.
\end{equation}
The epipolar distance for the trajectory $j$ is computed as the mean of all triplet epipolar distances along this trajectory. Concatenating all the trajectory epipolar distances gives a $1\times T$ matrix, $\mathbf{D}$.

We assign the epipolar distance of a trajectory to all the constituent pixels. Hence, the proposed approach can deal with parts of the foreground object that remain static for a few frames but were in motion otherwise. As we show in Fig.~\ref{fig:epipolarDistance} the epipolar distance estimated based on the current and the previous frame is quite small for the static part of the bear, whereas the trajectory-based epipolar distance can detect a significant part of the bear. Trajectory epipolar distances help us find powerful motion features for video segmentation, as we show in the next section.

\section{Approach}
\begin{figure*}    
\center
\rowSpace
\includegraphics[width=0.9\textwidth]{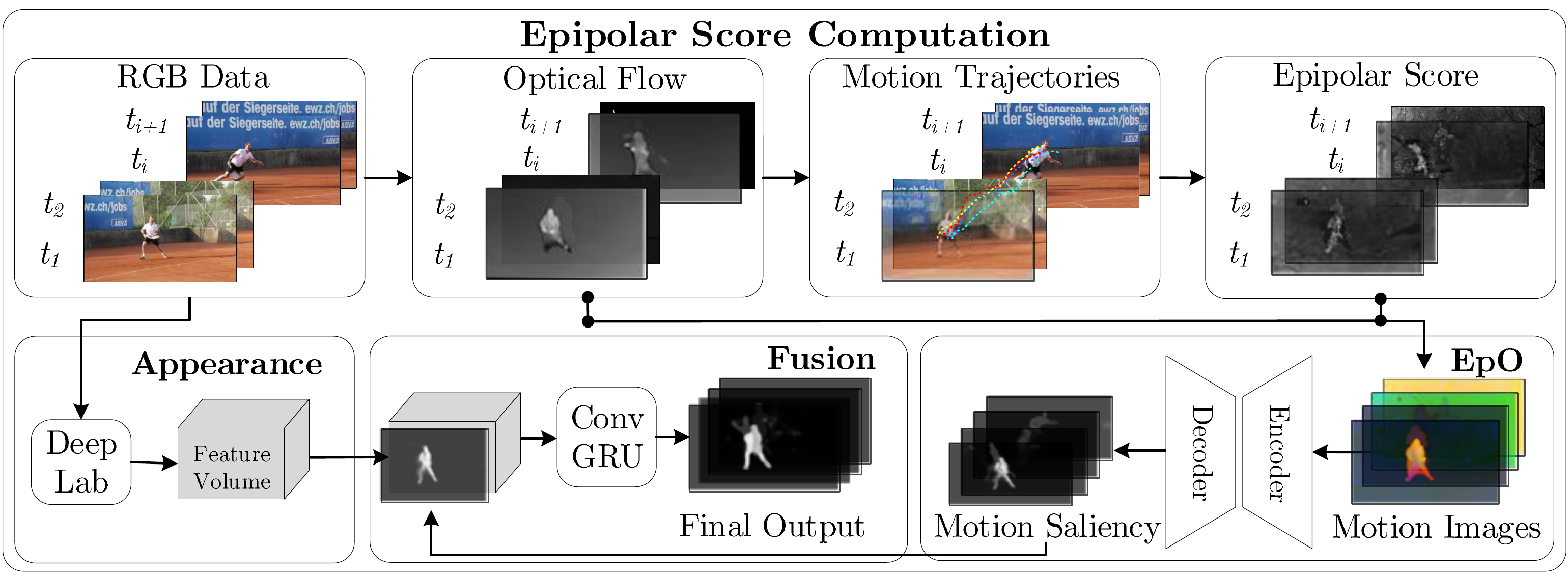}
% \vspace{-0.15in}
\caption{Flow diagram depicting different parts \& information transition in the algorithm. Top Row: steps to compute the motion trajectories \& Epipolar Distance. Bottom row: (Left) Deep-Lab based Appearance Network trained to compute the Appearance Features. (Right) Motion-Images (Optical Flow \& Epipolar Distance) fed to EpO, which outputs motion saliency map. (Middle) Motion-saliency map concatenated with appearance features are fed into the bidirectional convGRU.}
\rowSpace
\label{fig:systemDiag}
\end{figure*}

The proposed pipeline consists of three distinct stages as illustrated in Fig~\ref{fig:systemDiag}. 1) Our motion network, \textbf{EpO-Net} takes optical flow and epipolar distances as input, and outputs \textit{\textbf{motion-saliency-map}}. 2) Parallel to this, we have a network to compute the appearance features to extract scene context and object information \cite{deeplab}. 3) Our joint network, \textbf{EpO-Net+} fuses the appearance features and the motion-saliency-map with a bidirectional-ConvGRU and outputs saliency mask. %We introduce \textit{Input-Dropout}, a mechanism for robustly fusing noisy input feature-maps. We discuss these stages in detail as follows.

\subsection{Motion Images}
Given an input video, we compute optical flow, convert it into dense trajectories, find trajectory epipolar distances and convert them into per-frame \textit{Epipolar Distances} (ED).
Having a \textit{temporally} bigger receptive field, ED assigns a large weight to the foreground and lower to the background. 
% \textcolor{blue}{However, it} is sensitive towards optical flow errors, \textcolor{blue}{which it accumulates over the time }. 
However, it is sensitive towards optical flow errors because, during trajectory estimation, optical flow errors accumulate over time, affecting all the constituent pixels and their corresponding epipolar distances.
Whereas, optical flow captures temporally local but relatively robust information containing motion patterns to distinguish foreground from background. Both of them are complementary and should be exploited jointly. We concatenate 2-channel optical flow vectors with ED, to get a 3-channel image, we call \textit{motion-images}, as shown in Fig~\ref{fig:OurvsMpNet}.

%The main challenge in fusion is to identify when both of these information are reliable and when only optical flow should be used.

\subsection{Epipolar Optical flow Network (EpO-Net)}
Given the motion image as input, we design an encoder-decoder architecture, in the fashion of UNet \cite{uNet} that outputs motion-saliency-map.
The encoder latent space captures the context of the whole motion image, by jointly exploiting motion patterns and their relationship with ED. 
The decoding part on the other-hand has unraveled the context to decide about each pixel. 
The use of skip layers gives decoder access to local information (\cite{devildecoder}) collected from the lower layers of the encoding-network and helps to exploit the context to decide the pixel level labels. 

In our network, we use four encoders followed by four decoders, where each block consists of a convolution layer, followed by batch normalization, ReLU activation, and max-pooling layers. 
Different from Mp-Net, our much informative input allows us to have fewer channels before the final classification layer (128 instead of 512). Motion-saliency-map is produced using a sigmoid layer in the end. CRF is used to clean the output. A detailed architecture diagram showing the parameters of EpO $\&$ EpO$+$ is shown in the supplementary material.

%---------------------------------------------------------
\subsection{Joint Network (EpO-Net$+$)}

Any algorithm solely based on motion information will struggle with defining object boundaries and be confused by the non-rigid background. To exploit the additional appearance information, we use the pre-trained Deep-Lab~\cite{deeplab} features and fuse them with our motion network, similar to LVO \cite{visMem}. Although the FC6 layer of Deep-Lab is just $1/8$th of the spatial size of the original image, it still captures important information about the objects, their boundaries, and nonrigid background. Although customized appearance networks for video segmentation can produce better segmentation results, we choose to use rather generic appearance-based features, to demonstrate the significance of the proposed motion network.

We train the bottleneck layer to reduce the appearance features from 1024 to 128 and concatenate it with the down-sampled output of EpO-Net. To exploit temporal continuity in the joint-features and build a global context, we concatenate the bi-directional Convolutional Gated Recurrent Unit at the end of our network. To robustly handle motion network failures in the case of nonrigid background, we introduce input-dropout, discussed in Sec.~\ref{sec:inputDropOut}.

\section{Challenges in Training}
The proposed architecture contains a fusion of mixed features, encapsulating information at varied spatial and temporal receptive fields, at different stages of the network.
To enable the network to properly learn the concept of motion saliency, and robustly fuse these features, required contribution both in the training methodology and dataset.

\subsection{RBSF Dataset}
\label{sec:RBSFDataset}
%\section{REMOVE IT LATTER: Training EpO-Net}
Training sequences in the DAVIS 2016 are too few to train a  robust motion network. 
We find that F3DT \cite{F3DT} and PHAV \cite{PHAV} datasets are not very useful for us. F3DT has holes and the objects' motion is quite fast. PHAV is low resolution than DAVIS and the ground-truth optical flow is noisy because of jpeg compression.
We create our own synthetic dataset, called \textbf{RBSF} (Real Background, Synthetic Foreground), by mixing 20 different foreground objects performing various movements with 5 different real background videos.
With fairly large size objects (size: 30\% to 50\% of the frame) and reasonably fast motion, RBSF allows us to compute accurate optical flow and long trajectories. We observe that generating more data does not improve results, thanks to the well-constrained epipolar distances. After training on RBSF, we fine-tune EpO-Net on DAVIS-2016 \cite{DAVIS2016}. For more details of the dataset, please see the supplementary material.

\subsection{Feature Fusion \& Input-Dropout}
\label{sec:inputDropOut}
Robustly fusing optical flow and epipolar distances is a challenging task. Ideally, the network should be able to learn which feature to rely on the pixel-level granularity. But this requires contextual information that is only available in the deeper layers of the network, where the resolution is usually very small and the network has already mixed the input channels. In such a scenario, training with more data or for more iterations usually does not improve the results.

This problem is usually solved by introducing a mix of early and late fusion, requiring complex network designs, where skip layers are going from one part of the network to the others. Our proposed solution is rather quite simple, which we call \textbf{Input-Dropout}. 
% \textcolor{blue}{
% It entails enabling the network to learn how to mix the features by withholding the complete information or replacing it with the noisy information at the input level. 
% Since, CNN consists of multiple filters, some of the filters learn at that layer to ignore parts of the input others learn to mix them. This is unlike the simple dropout method, where ... I AM STILL WORKING ON HOW TO DIFFERENTIATE IT WITH SIMPLE DROPOUT LAYER. }
While training EpO-Net, we randomly make complete ED-channel zero, for some of the sequences which have erroneous ED-maps (sequences with large motion and a considerable occlusion). For the rest, motion-images are unaltered. This is done for the initial 10 epochs, allowing the filters to give more importance to the optical flow. After that, we repeat the same procedure but instead of zero, we assign random values, forcing the network to learn the diverse enough filters to capture the motion information from the optical flow, ED and their combination, separately. With input-dropout EpO's mean IoU increases from $72.7$ to $75.2$ (Table~\ref{tab:ablation}).

We exploit the same input-dropout strategy for the late fusion of appearance and motion features in our joint network. 
%Training Fusion-Net required the same input-dropout strategies, as used in EpO-Net, to get robust joint features.
We randomly set the motion-saliency-map to zero for a few frames of the sequences, where the motion network fails (sequences with dynamic background and occlusion). Using input-dropout, mean IoU improves from $79.4$ to $80.6$.
The complete network, containing all the above stages and layers is called \textbf{EpO-Net+}.
% \rowSpace
\vspace{-0.20cm}

\section{Experiments}

\setlength{\tabcolsep}{0.5pt}
\begin{figure*}[!th]    
\center
\includegraphics[width=1.0\textwidth]{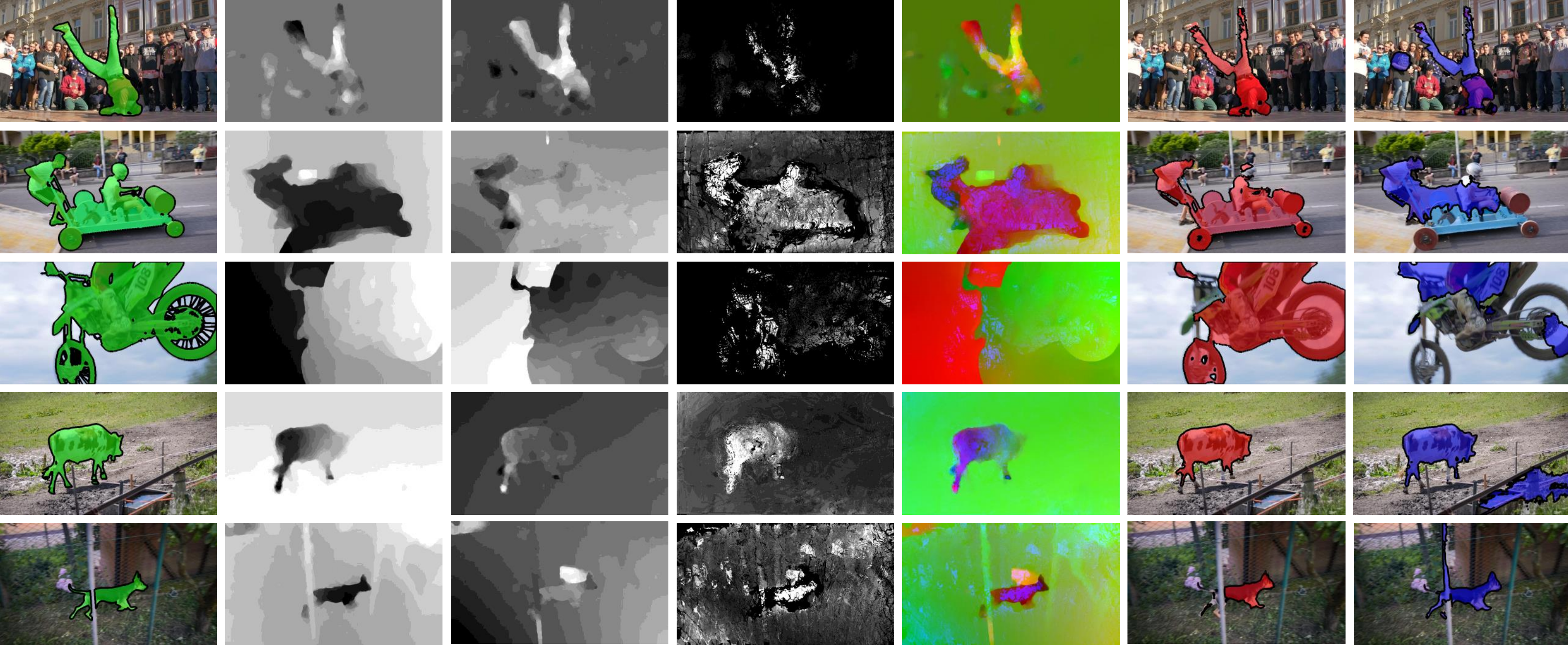}

\begin{tabular}{P{1.0in}P{0.9in}P{1.0in}P{0.95in}P{0.95in}P{0.95in}P{1.0in}}

\small{Ground truth} & \small{X-Displancement} & \small{Y-Displacement} & \small{ED}  & \small{Motion Images} & \small{EpO-Net} & \small{Mp-Net~\cite{MpNet}} 
\end{tabular}
\rowSpace
\caption{Qualitative Comparison of our EpO-Net with Mp-Net~\cite{MpNet}.}
\label{fig:OurvsMpNet}
\rowSpace
\end{figure*}

%%%%%%%%%%%%%%%%%%%%%%%%%%%%%%%%%%%%%%%%%%%%%%%%%%%%%%%%%
We train and evaluate on RBSF (Sec. \ref{sec:RBSFDataset}), DAVIS2016 \cite{DAVIS2016}, DAVIS2017 \cite{DAVIS2017} and Segtrack-v2  \cite{segtrackv2}. 
Below we detail our training parameters and evaluation results. 

\setlength{\tabcolsep}{1.0pt}
\begin{table}
\begin{center}
\centering
% \label{Epo-vs-MPNet}
\begin{tabular}{c|cccccc}
\hline
\hline
Method & AC & DB & FM & MB & OCC & Mean \\
\hline
Mp-Net & 0.71 \emph{\scriptsize{-0.02}} & 0.58 \emph{\scriptsize{0.14}} & 0.68 \emph{\scriptsize{0.04}} &  0.65 \emph{\scriptsize{0.10}} &\bf 0.69 \emph{\scriptsize{0.01}} & 0.700 \\

EpO & \bf 0.77 \emph{\scriptsize{-0.03}} & \bf 0.63 \emph{\scriptsize{0.14}} & \bf 0.72 \emph{\scriptsize{0.06}} & \bf 0.67 \emph{\scriptsize{0.14}} & 0.67 \emph{\scriptsize{0.11}} & \bf\ 0.752 \\
\hline
\hline
\end{tabular}
\caption{EpO-Net vs. Mp-Net \cite{MpNet} on DAVIS-2016 dataset. \label{tab:EpoVsMP}}
\rowSpace
\rowSpace
\rowSpace
\end{center}
\end{table}
%%%%%%%%%%%%%%%%%%%%%%%%%%%%%%%%%%%%%%%%%%%%%%%%%%%%%%%%%%%%%%%%%%%%%%%%%%%%

\subsection{Implementation Details}
\label{implementationDetails}
  
EpO is trained using the mini-batch SGD with a batch size of 12, the initial learning rate is set to 0.001, with a momentum of 0.9, and a weight decay of 0.005. 
The network is trained from scratch for 50 epochs, with the learning rate decay factor set to 0.1, after every 5 epochs. 
The images are down-sampled by a factor of 0.5 to fit a batch size of 12 images in the GPU memory.
  
We train EpO in two stages: training on a synthetic dataset, RBSF (Sec. \ref{sec:RBSFDataset}), and then fine-tuning on DAVIS-2016. For both of these training, we perform input-dropout for epipolar channel for only 20\% of training data i.e. we randomly assign zero and add small random Gaussian noise in the epipolar channel. We call this final trained model, \textbf{EpO} and the one trained on RBSF, \textbf{EpO-RBSF}.

The fusion network is fully trained only on the DAVIS-2016's training set, resulting in \textbf{EpO+}.
We use the batch size of 12 and an initial learning rate set to 0.001, which is decreased after every epoch with a factor \(\frac{epoch}{50}\). 
The model is trained using the back-propagation through time \cite{backpropagationThrougTime} using binary cross-entropy loss and RMSProp optimizer. 
The weights of all the layers in the fusion network are initialized using the Xavier \cite{xavier}, except for those in ConvGRU, that is initialized using MSR initialization~\cite{MSR}. We clip the gradients to the [-50, 50], before each update step \cite{graves2013generating} to avoid numerical issues. 
For robust fusion, we again use the input-dropout mechanism by setting the motion-saliency-map to zero, for 20\% frames of the sequence with fast motion and dynamic background. We also perform the random cropping and flipping of sequences during the training. The fusion network is trained for 50 epochs. The final output is refined using CRF, during inference. 
To test on DAVIS-2017, we fine-tine EpO-RBSF and EpO on the DAVIS-2017's training-set.

\vspace{0.125cm}

%UPDATED TABLE
%%%%%%%%%%%%%%%%%%%%%%%%%%%%%%%%%%%%%%%%%%%%%%%%%%%%%%%%%%%%%%%%%%%%%%%%%%%%
\setlength{\tabcolsep}{1.4pt}
\begin{table*}[!h]
\begin{center}
\centering
\begin{tabular}{cc|cc|cccccccccc}
\hline
\hline

\multicolumn{2}{c|}{Measure} & EpO+ & EpO & \small{AGS\cite{ags}} & \small{MOA\cite{MotAdapt}} & \small{LSMO\cite{lsmo}}  & \small{STP\cite{STP}} &  \small{PDB\cite{PDB}} & \small{ARP\cite{ARP}} & \small{LVO\cite{visMem}} & \small{Mp-Net\cite{MpNet}} & \small{FSeg\cite{fusionseg}} & \small{SFL\cite{SegFlow}}   \\  
\hline

&Mean $\mathcal{M} \uparrow$ &\bf\ 0.806 & \ 0.752 & \ \underline{0.797} & \  0.772    & \  0.782     & \ 0.776 & \ 0.772 & \ 0.762 & 0.759 & \ 0.700 & \ 0.707 & \ 0.674  \\

$\mathcal{J}$ & Recall $\mathcal{O} \uparrow$   & \bf \ 0.952 & \ 0.888 & \  \underline{0.911}  & \ 0.878     & \ 0.891 & \ 0.886 &  \ 0.901 &   \ 0.911 & \ 0.891 &   \ 0.850 &   \ 0.835 &  \ 0.814  \\

&Decay $\mathcal{D} \downarrow$  &\ 0.022 & \ 0.053 & \ 0.019  & \ 0.050    & \ 0.041     & \ 0.044 & \ \underline{0.009} & \ 0.070 &\bf\ 0.000 &   \ 0.013 &   \ 0.015 & \ 0.062 \\
\hline
& Mean $\mathcal{M} \uparrow$ & \  \underline{0.755} & \ 0.711 & \bf 0.774  & \ 0.774 & \ 0.759 & \ 0.750 &  \ 0.745 &   \ 0.706 & \ 0.721 &   \ 0.659 &   \ 0.653 &   \ 0.667 \\
$\mathcal{F}$ & Recall $\mathcal{O} \uparrow$   &\bf\ 0.879 & \ 0.830 & \ \underline{0.858}  & \ 0.844    & \  0.847  & \ 0.869 & \ 0.844 &   \ 0.835 & \ 0.834 &   \ 0.792  &   \ 0.738 &   \ 0.771  \\
& Decay $\mathcal{D} \downarrow$  & \ 0.024 & \ 0.043  & \ 0.016  & \ 0.033 & \     0.035    & \ 0.042 &\bf\ -0.002 &   \ 0.079 & \ \underline{0.013} &   \ 0.025 &   \ 0.018  &   \ 0.051 \\
\hline
$\mathcal{T}$ & Mean $\mathcal{M} \downarrow$ &\bf\ 0.185 & \ 0.388 & \ 0.267  & \ 0.279     & \  0.212    & \ \underline{0.243} &  \ 0.277 &   \ 0.384 &   \ 0.255 &   \ 0.563 &   \ 0.316 &   \ 0.282  \\
\hline
\hline
\end{tabular}
\vspace{-0.3cm}
\caption{Comparison of our motion (EpO) and fusion network (EpO+), with state-of-the-art on DAVIS-2016 with intersection over union $\mathcal{J}$, F-measure $\mathcal{F}$, and temporal stability $\mathcal{T}$. Best \& second best scores have been bold and are underlined respectively. AGS uses eye-gaze data to train their network, whereas we only exploit information existent in the videos itself by enforcing the geomatrical constraints.\label{tab:comparisonTable2016}} 
\vspace{-0.3in}
\end{center}
\end{table*}
%\setlength{\tabcolsep}{1.4pt}

%%%%%%%%%%%%%%%%%%%%%%%%%%%%%%%%%%%%%%%%%%%%%%%%%%%%%%%%%%%%%%%%%%%%%%%%%%%%

%%%%%%%%%%%%%%%%%%%%%%%%%%%%%%%%%%%%%%%%%%%%%%%%%%%%%%%%%%%%%%%%%%%%%%
%ATTRIBUTES

\setlength{\tabcolsep}{1.4pt}
\begin{table}[!h]
\begin{center}
\centering
\begin{tabular}{c|c|cccc}
\hline
\hline
Attribute & EpO+ & AGS\cite{ags} & MOA\cite{MotAdapt} & LSMO\cite{lsmo} & STP\cite{STP} 
\\
%& FST\cite{FST} &  NLC\cite{NLC} \\ 
% EpO & \bf 0.77 \emph{\scriptsize{-0.03}} & \bf 0.63 \emph{\scriptsize{0.14}} & \bf 0.72 \emph{\scriptsize{0.06}} & \bf 0.67 \emph{\scriptsize{0.14}} & 0.67 \emph{\scriptsize{0.11}} & \bf\ 0.752 \\
\hline
AC    
&\bf 0.83 \emph{\scriptsize{-0.04}}     & \underline{0.80
\emph{\scriptsize{-0.01}}}    &    0.78
\emph{\scriptsize{-0.01}}    &    0.78
\emph{\scriptsize{+0.00}}    &    0.72 \emph{\scriptsize{+0.07}}
\\
DB    &\bf 0.72 \emph{\scriptsize{+0.10}}    & \underline{0.66
\emph{\scriptsize{+0.16}}}    & 0.61
\emph{\scriptsize{+0.20}}    & 0.55
\emph{\scriptsize{+0.27}}    & 0.66 \emph{\scriptsize{+0.15}}
\\
FM    &\bf 0.78 \emph{\scriptsize{+0.04}}     & \underline{0.77
\emph{\scriptsize{+0.04}}}    & 0.74
\emph{\scriptsize{+0.05}}    & 0.73
\emph{\scriptsize{+0.08}}    & 0.75 \emph{\scriptsize{+0.04}}
\\
MB    &\bf 0.78 \emph{\scriptsize{+0.06}}    & \underline{0.74 \emph{\scriptsize{+0.10}}}    &    0.71 \emph{\scriptsize{+0.10}}    &    0.73 \emph{\scriptsize{+0.10}}    &    0.74 \emph{\scriptsize{+0.06}}          
\\
OCC    & 0.75 \emph{\scriptsize{+0.08}} &    0.76 \emph{\scriptsize{+0.05}}    & \underline{0.78 \emph{\scriptsize{-0.02}}}    &    0.74 \emph{\scriptsize{+0.06}}    &   \bf 0.81
\emph{\scriptsize{-0.05}}
\\
\hline
\hline
\end{tabular}
\caption{Attribute-based analysis of top performing methods on DAVIS-2016 dataset. The mean IoU on all sequences with attributes: appearance cahnge (AC), dynamic background (DB), fast motion (FM), motion blur (MB), and occlusion (OCC), is computed. The smaller font values indicate the change in performance (gain or loss) for the method on the remaining sequences without that respective attribute. \label{tab:attributeAnalysisDAVIS2016}}
\rowSpace
\rowSpace
\end{center}
\end{table}
\setlength{\tabcolsep}{1.4pt}
\vspace{-0.1cm}

%%%%%%%%%%%%%%%%%%%%%%%%%%%%%%%%%%%%%%%%%%%%%%%%%%%%%%%%%%%%%%%%%%%%%%
\subsection{Evaluation}
We follow the standard training \& validation split, to train and evaluate using the protocol proposed in \cite{DAVIS2016} and compute intersection-over-union $\mathcal{J}$, F-measures $\mathcal{F}$, and temporal stability $\mathcal{T}$ 
%for measuring region similarity
, contour accuracy and smoothness of segmentation overtime respectively. The evaluation results are summarized in Table~\ref{tab:comparisonTable2016}.

\vspace{-0.25cm}

\subsubsection{Motion Network} 
By exploiting geometric constraints in salient motion segmentation, our EpO (motion-only) network scores mean $\mathcal{J}$ of $0.752$ over \textbf{DAVIS-2016 validation set}. 
This is much higher than $0.70$ score of Mp-Net~\cite{MpNet}, which also relies on non-motion features (objectness score). MP-Net is trained on 45K frames using ground-truth optical flow, whereas EpO uses only 20K frames and an estimated optical flow on them. We observe that using more data does not improve the performance, thanks to the well-constrained epipolar distances. Moreover, our EpO score is competitive to LVO, which is using a bi-directional ConvGRU and the appearance information in addition to optical flow. Whereas EpO only uses motion-images (optical flow \& ED). 

\textbf{Qualitative comparison} of EpO-Net with Mp-Net is given in Fig.~\ref{fig:OurvsMpNet}. 
It's evident from the $2^{nd}$ to $4^{th}$ columns that ED and optical flow are complimenting each other, and the results are robust against the failure of one of these inputs.
In the case of optical flow being too small, or if the object motion is in the same direction as the camera motion (row-2), ED helps distinguish the object. 
Similarly, when the ED score is sporadically bad (row-1 \& 3), optical flow information helps to distinguish the object, much due to the robust motion features learned with input-dropout training.
Whereas Mp-Net makes local decisions, unable to recover from the optical flow errors (row 3 \& 5). Their results also degrade when the camera and object have similar motion (row-3).  
\rowSpace

%%%%%%%%%%%%% END OF MOTION NETWORK
\subsubsection{EpO+}
Combining the motion-saliency map obtained from EpO with the appearance features and adding temporal memory, EpO+ outperforms its direct competitors LVO and LSMO, by a significant margin of 4.7\% and 2.4\% over mean IoU. 
EpO+ outperforms even recently published works, like AGS \cite{ags}, which requires training on dynamic fixation dataset collected by tracking the gaze of viewers, both in mean IoU and its recall. 
Important to note is mean temporal stability, which is substantially better than rest explicitly indicating the effectiveness of our formulation. 
Our attribute analysis is given in Table~\ref{tab:attributeAnalysisDAVIS2016}. EpO+ outperforms the baselines in all categories except the occlusion.

\begin{figure*}    
\center
\includegraphics[width=1.0\textwidth]{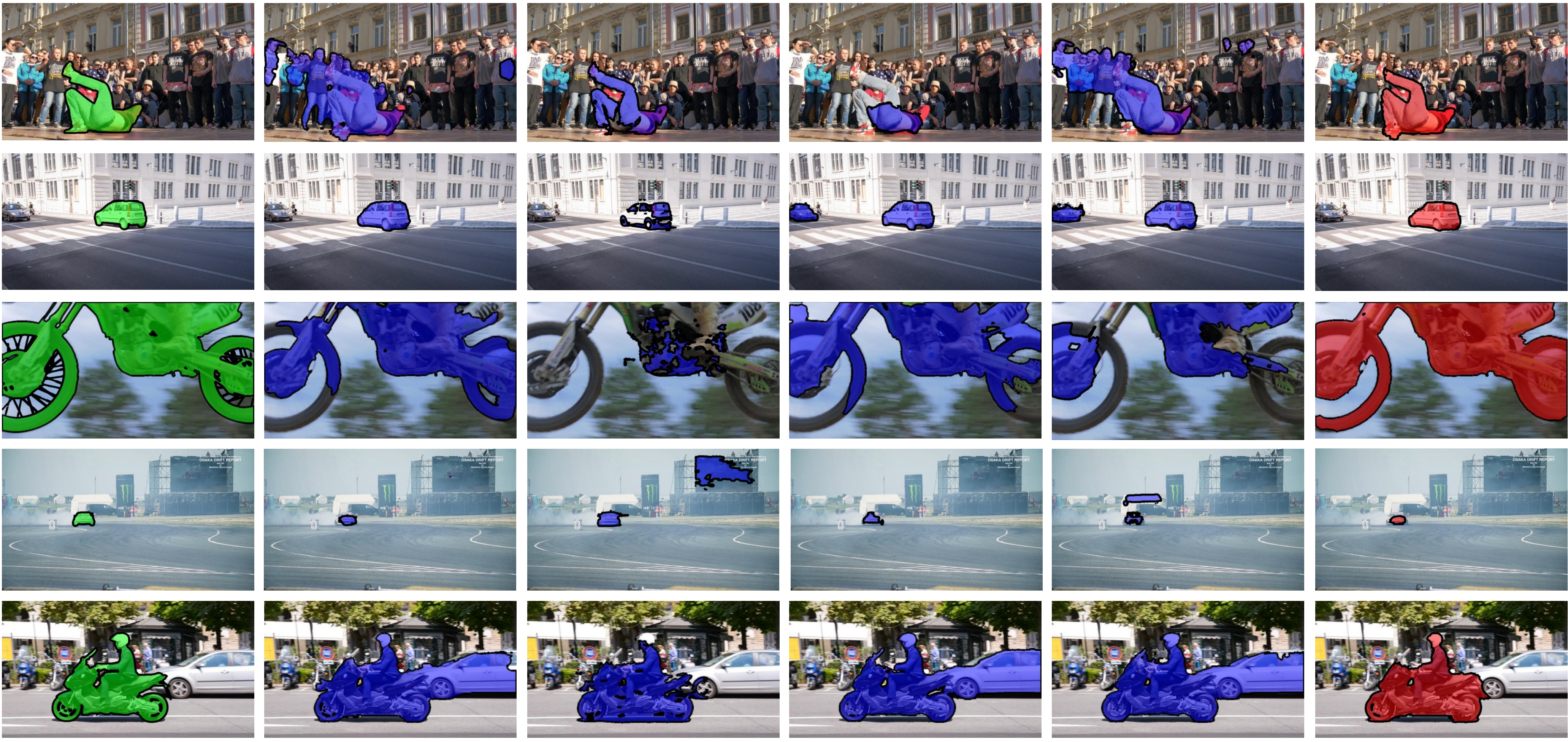}

\begin{tabular}{P{1.0in}P{1.2in}P{1.2in}P{1.05in}P{1.1in}P{1.0in}}
\small{Ground truth} & \small{LVO~\cite{visMem}} & \small{STP~\cite{STP}} & \small{MotAdapt~\cite{MotAdapt}} & \small{AGS~\cite{ags}} & \small{Our}
\\
\end{tabular}
\vspace{-0.1in}
\caption{Qualitative comparison with state-of-the-art methods on DAVIS-2016.}
\vspace{-0.2in}

\label{tab:qualitativeResults}
\end{figure*}

\textbf{Qualitative comparison} of EpO+ with SOTA is presented in Fig.~\ref{tab:qualitativeResults}.
AGS has failed to properly segment moving objects ($2^{\textrm{nd}}$ and $3^\textrm{rd}$ row). Most of the errors in the previous methods are over-segmenting and are due to over-exploitation of appearance information. 
%The fundamental reason behind this is that these methods are unable to exploit/learn enough constraints for motion saliency.
This we can attribute to the very basic reason of not being able to exploit/learn enough constraints for motion saliency.

While the proposed method, due to more informative proposed motion features (based on geometric constraints) and input-dropout training procedure, is being able to learn how to balance appearance and motion cues. For details see supplementary material. 
\rowSpace

\subsubsection{Evaluation on other datasets}
\textbf{DAVIS-2017:} We fine-tune EpO-RBSF and EpO+ on the DAVIS-2017's training sequences. We could not find the comparative results, but we are reporting ours for future comparison in Table~\ref{Tab:DAVIS2017}.  

\textbf{Segtrack-v2:} Evalaution resutls of EpO+ and EpO on SegTrack-v2 \cite{segtrackv2} dataset have been presented in Table ~\ref{tab:quantSegTrack}. 
Our results are better than existing methods, including STP \cite{STP}.
Although, it's with a small margin of $0.8\%$; this could be attributed to the difference in resolution of SegTrack-v2 videos vs that of DAVIS-2016. Removing birdfall, the only sequence we perform poorly, the results improves to $72.8\%$. AGS~\cite{ags} uses both SegTrackv2 and DAVIS for training, therefore, do not evaluate on this. Note that, since NLC  \cite{NLC} reports results only on subset of sequences in their paper, results in Table ~\ref{tab:quantSegTrack} were taken from \cite{visMem, STP}.

\setlength{\tabcolsep}{1pt}

\setlength{\tabcolsep}{1.6pt}
\begin{table}
\begin{center}
\centering
\begin{tabular}{l|cccccccc}
\hline
\hline
Method & KEY & NLC & FSG & LVO & LSMO & STP & EpO & EpO+ \\
\hline
Mean IoU & 57.3 & 67.2 & 61.4 & 57.3 & 59.1 & \underline{70.1} & 68.3 & \bf{70.9} \\
\hline
\hline
\end{tabular}
\caption{EpO+ results on SegTrack-v2 dataset \cite{segtrackv2}. We only perform bad on one sequence (birdfall). Removing this increase our Mean IoU to 72.8. \label{tab:quantSegTrack}
}
\vspace{-0.3in}
\end{center}
\end{table}

\setlength{\tabcolsep}{1.0pt}
\begin{table}[!h]
\begin{center}
\centering
\begin{tabular}{c|cccccc}
\hline
\hline
Method & AC & DB & FM & MB & OCC & $\mathcal{J}$ Mean \\  
\hline
EpO & 0.67 \emph{\scriptsize{-0.02}} & 0.56 \emph{\scriptsize{0.10}} & 0.62 \emph{\scriptsize{0.04}} &  0.57 \emph{\scriptsize{0.11}} & 0.59 \emph{\scriptsize{0.08}} & 0.652 \\

EpO+ & 0.79 \emph{\scriptsize{-0.04}} & 0.72 \emph{\scriptsize{0.05}} & 0.74 \emph{\scriptsize{0.03}} &  0.72 \emph{\scriptsize{0.06}} & 0.66 \emph{\scriptsize{0.13}} & 0.763 \\
\hline
\hline
\end{tabular}
\caption{Results on DAVIS 2017 dataset.\label{Tab:DAVIS2017}}
\rowSpace
\rowSpace
\end{center}
\end{table}

%%%%%%%%%%%%%%%%%%%%%%%%%%%%%%%%%% End Evaluation    %%%%%%%%%%%%%%%%%%%%%%%%%%%%%%%%%%%%%%%%

\setlength{\tabcolsep}{1.4pt}
\begin{table}
\begin{center}
\centering
\begin{tabular}{c|ccc}
    \hline
    \hline
    {\#enc/dec} & \multicolumn{3}{c}{Input Modality} \\
     & Ep & OF & Ep+OF \\
    \hline
    2 & 57.2 & 54.7 & 62.7 \\
    % \hline
    3 & 58.9 & 59.7 & 64.4 \\
    % \hline
    4 & 49.2 & 63.3 & 67.5 \\
    \hline
\end{tabular}
\quad
\begin{tabular}{l|c}
    \hline
    \hline
    EpO Variant & Mean IoU \\
    \hline
    EpO{\scriptsize{(R)}} & 48.5 \\
    EpO{\scriptsize{(D)}} & 72.7 \\
    EpO{\scriptsize{(R)}}+Drop & 50.6 \\
    EpO{\scriptsize{(D)}}+Drop & 75.2 \\
    \hline
\end{tabular}
\end{center}
\vspace{-0.2in}
\caption{\textbf{Left:} Studying the effects of different input modalities against network depth. \textbf{Right:} Effect of dropout in epipolar channel of motion images, R and D denote RBSF and DAVIS dataset respectively. \label{tab:ablation}} \rowSpace 
\end{table}
%%%%%%%%%%%%%%%%%%%%%%%%%%%%%%%%%%%%   Start Ablation  %%%%%%%%%%%%%%%%%%%%%%%%%%%%%%%%
\subsection{Ablation Study}
In this section, we present the study on the impact and effectiveness of different design choices. We first analyze the influence of different input modalities and depth of the network architecture by training and validating on DAVIS-2016 dataset. Specifically, we use the single-channel ED, 2 channel optical flow i.e. X-Y displacement, and the combination of the both as 3 channel \textit{motion images}. For each input modality, we train and validate EpO network with 2, 3 and 4-layer encoders/decoders to study which modality needs the deeper network.

In Table~\ref{tab:ablation}, we observe that ED being a very simple yet informative feature, the epipolar alone network requires fewer parameters to learn, implying that they should not require (i) deep network, ii) large datasets. In contrast, optical flow, being complex information for motion saliency, requires more encoders and decoders. Since small errors in optical flow, get accumulated in trajectories estimation and result in quite noisy epipolar distances, optical flow with 4 encoders/decoders architecture beats the epipolar network, with 63.3\% mean IoU. However, when we combine both, in the form of motion images, the accuracy further improves by 4.2\%. This shows that the combination can exploit both the global temporal geometric information and local temporal motion information distinguishing the foreground and background. Note that all the experiments are performed using the same hyper-parameters stated in Sec.~\ref{implementationDetails}, the input-dropout strategy is not used, and all models are trained for 30 epochs only.

Next, we demonstrate the effectiveness of our dataset RBSF and the input-dropout in Table~\ref{tab:ablation}. The mean IoU on DAVIS-2016 with the proposed dataset was 48.5\%. That increases to 72.7\% with fine-tuning on DAVIS. Comparing this with our Ep+OF's, the increase is 5.3\%, showing the significance of the proposed dataset. With the proposed dropout the results further improve by 2.5\%, showing the effectiveness of the input-dropout. 

We also study the effect of GRU-sequence length. As expected, when we increase sequence length, from 6 to 12, the mean IoU improves from $77.3$ to $79.4$. A considerable improvement comes in the videos having occlusion. Finally, we observe that instead of the angle-magnitude representation of optical flow, the velocity representation gives better results. A qualitative review of the dataset, made us realize that the channel representing angle information is not robust to optical flow errors. Even for humans, inferring motion patterns by just looking at them, is quite difficult.

\section{Conclusion}
We exploit multiview geometric constraints to define motion saliency and propose trajectory epipolar distances, as a measure of non-rigidity. By combining epipolar distances with optical flow, we train a powerful motion network and demonstrate significant improvement over the previous motion networks. Unlike previous methods, the learned motion features avoid over-reliance on appearance-based features. Even without using RNNs and appearance features, our motion network is competitive to the existing state of the art. With them, our method gives state of the art results.
An input-dropout mechanism has been proposed that allows network to learn robust feature fusion. 
The proposed learning paradigm, involving the strong geometric constraints, should be useful for a number of related applications.  
%The proposed input-dropout idea may also be useful to learn robust joint features in network fusion.

{\small
\bibliographystyle{ieee}
\bibliography{egbib}

\begin{thebibliography}{10}\itemsep=-1pt

\bibitem{CNN_MRF}
L.~Bao, B.~Wu, and W.~Liu.
\newblock Cnn in mrf: Video object segmentation via inference in a cnn-based
  higher-order spatio-temporal mrf.
\newblock In {\em Proceedings of the IEEE Conference on Computer Vision and
  Pattern Recognition}, pages 5977--5986, 2018.

\bibitem{brox2010object}
T.~Brox and J.~Malik.
\newblock Object segmentation by long term analysis of point trajectories.
\newblock In {\em European conference on computer vision}, pages 282--295.
  Springer, 2010.

\bibitem{deeplab}
L.~Chen, G.~Papandreou, I.~Kokkinos, K.~Murphy, and A.~L. Yuille.
\newblock Semantic image segmentation with deep convolutional nets and fully
  connected crfs.
\newblock In {\em 3rd International Conference on Learning Representations,
  {ICLR} 2015, San Diego, CA, USA, May 7-9, 2015, Conference Track
  Proceedings}, 2015.

\bibitem{chen2016full}
Q.~Chen and V.~Koltun.
\newblock Full flow: Optical flow estimation by global optimization over
  regular grids.
\newblock In {\em Proceedings of the IEEE Conference on Computer Vision and
  Pattern Recognition}, pages 4706--4714, 2016.

\bibitem{ChenPMG18}
Y.~Chen, J.~Pont{-}Tuset, A.~Montes, and L.~V. Gool.
\newblock Blazingly fast video object segmentation with pixel-wise metric
  learning.
\newblock In {\em 2018 {IEEE} Conference on Computer Vision and Pattern
  Recognition, {CVPR} 2018, Salt Lake City, UT, USA, June 18-22, 2018}, pages
  1189--1198, 2018.

\bibitem{SegFlow}
J.~Cheng, Y.-H. Tsai, S.~Wang, and M.-H. Yang.
\newblock Segflow: Joint learning for video object segmentation and optical
  flow.
\newblock In {\em 2017 IEEE International Conference on Computer Vision
  (ICCV)}, pages 686--695. IEEE, 2017.

\bibitem{PHAV}
C.~De~Souza, A.~Gaidon, Y.~Cabon, and A.~Lopez~Pena.
\newblock Procedural generation of videos to train deep action recognition
  networks.
\newblock In {\em CVPR}, 2017.

\bibitem{endres2010category}
I.~Endres and D.~Hoiem.
\newblock Category independent object proposals.
\newblock In {\em European Conference on Computer Vision}, pages 575--588.
  Springer, 2010.

\bibitem{NLC}
A.~Faktor and M.~Irani.
\newblock Video segmentation by non-local consensus voting.
\newblock In {\em BMVC}, volume~2, page~8, 2014.

\bibitem{feng2004multiple}
S.~Feng, R.~Manmatha, and V.~Lavrenko.
\newblock Multiple bernoulli relevance models for image and video annotation.
\newblock In {\em null}, pages 1002--1009. IEEE, 2004.

\bibitem{fragkiadaki2012video}
K.~Fragkiadaki, G.~Zhang, and J.~Shi.
\newblock Video segmentation by tracing discontinuities in a trajectory
  embedding.
\newblock In {\em Computer Vision and Pattern Recognition (CVPR), 2012 IEEE
  Conference on}, pages 1846--1853. IEEE, 2012.

\bibitem{xavier}
X.~Glorot and Y.~Bengio.
\newblock Understanding the difficulty of training deep feedforward neural
  networks.
\newblock In {\em Proceedings of the Thirteenth International Conference on
  Artificial Intelligence and Statistics}, pages 249--256, 2010.

\bibitem{graves2013generating}
A.~Graves.
\newblock Generating sequences with recurrent neural networks.
\newblock {\em arXiv preprint arXiv:1308.0850}, 2013.

\bibitem{hartley2003multiple}
R.~Hartley and A.~Zisserman.
\newblock {\em Multiple view geometry in computer vision}.
\newblock Cambridge university press, 2003.

\bibitem{MSR}
K.~He, X.~Zhang, S.~Ren, and J.~Sun.
\newblock Delving deep into rectifiers: Surpassing human-level performance on
  imagenet classification.
\newblock In {\em 2015 {IEEE} International Conference on Computer Vision,
  {ICCV} 2015, Santiago, Chile, December 7-13, 2015}, pages 1026--1034, 2015.

\bibitem{videoMatchECCV2018}
Y.~Hu, J.~Huang, and A.~G. Schwing.
\newblock Videomatch: Matching based video object segmentation.
\newblock In {\em Computer Vision - {ECCV} 2018 - 15th European Conference,
  Munich, Germany, September 8-14, 2018, Proceedings, Part {VIII}}, pages
  56--73, 2018.

\bibitem{STP}
Y.-T. Hu, J.-B. Huang, and A.~Schwing.
\newblock Unsupervised video object segmentation using motion saliency-guided
  spatio-temporal propagation.
\newblock In {\em Proc. ECCV}, 2018.

\bibitem{fusionseg}
S.~D. Jain, B.~Xiong, and K.~Grauman.
\newblock Fusionseg: Learning to combine motion and appearance for fully
  automatic segmention of generic objects in videos.
\newblock {\em arXiv preprint arXiv:1701.05384}, 2(3):6, 2017.

\bibitem{cliqu2018}
Y.~Jun~Koh, Y.-Y. Lee, and C.-S. Kim.
\newblock Sequential clique optimization for video object segmentation.
\newblock In {\em Proceedings of the European Conference on Computer Vision
  (ECCV)}, pages 517--533, 2018.

\bibitem{CUT}
M.~Keuper, B.~Andres, and T.~Brox.
\newblock Motion trajectory segmentation via minimum cost multicuts.
\newblock In {\em Computer Vision (ICCV), 2015 IEEE International Conference
  on}, pages 3271--3279. IEEE, 2015.

\bibitem{ARP}
Y.~J. Koh and C.-S. Kim.
\newblock Primary object segmentation in videos based on region augmentation
  and reduction.
\newblock In {\em Proceedings of the IEEE Conference on Computer Vision and
  Pattern Recognition}, volume~1, page~6, 2017.

\bibitem{KEY}
Y.~J. Lee, J.~Kim, and K.~Grauman.
\newblock Key-segments for video object segmentation.
\newblock In {\em IEEE International Conference on Computer Vision}, pages
  1995--2002. IEEE, 2011.

\bibitem{segtrackv2}
F.~Li, T.~Kim, A.~Humayun, D.~Tsai, and J.~M. Rehg.
\newblock Video segmentation by tracking many figure-ground segments.
\newblock In {\em Proceedings of the IEEE International Conference on Computer
  Vision}, pages 2192--2199, 2013.

\bibitem{instanceEmbTrans2018}
S.~Li, B.~Seybold, A.~Vorobyov, A.~Fathi, Q.~Huang, and C.-C.~J. Kuo.
\newblock Instance embedding transfer to unsupervised video object
  segmentation.
\newblock In {\em Proceedings of the IEEE Conference on Computer Vision and
  Pattern Recognition}, pages 6526--6535, 2018.

\bibitem{PReMVOS}
J.~Luiten, P.~Voigtlaender, and B.~Leibe.
\newblock Premvos: Proposal-generation, refinement and merging for video object
  segmentation.
\newblock In {\em 14th Asian Conference on Computer Vision, Perth, Australia,
  December 2-6, 2018, pages = {565--580}}.

\bibitem{ma2012maximum}
T.~Ma and L.~J. Latecki.
\newblock Maximum weight cliques with mutex constraints for video object
  segmentation.
\newblock In {\em Computer Vision and Pattern Recognition (CVPR), 2012 IEEE
  Conference on}, pages 670--677. IEEE, 2012.

\bibitem{osvosJournal}
K.~Maninis, S.~Caelles, Y.~Chen, J.~Pont{-}Tuset, L.~Leal{-}Taix{\'{e}},
  D.~Cremers, and L.~V. Gool.
\newblock Video object segmentation without temporal information.
\newblock {\em {IEEE} Trans. Pattern Anal. Mach. Intell.}, 41(6):1515--1530,
  2019.

\bibitem{F3DT}
N.~Mayer, E.~Ilg, P.~Hausser, P.~Fischer, D.~Cremers, A.~Dosovitskiy, and
  T.~Brox.
\newblock A large dataset to train convolutional networks for disparity,
  optical flow, and scene flow estimation.
\newblock In {\em Proceedings of the IEEE Conference on Computer Vision and
  Pattern Recognition}, pages 4040--4048, 2016.

\bibitem{NagarajaSB15}
N.~S. Nagaraja, F.~R. Schmidt, and T.~Brox.
\newblock Video segmentation with just a few strokes.
\newblock In {\em 2015 {IEEE} International Conference on Computer Vision,
  {ICCV} 2015, Santiago, Chile, December 7-13, 2015}, pages 3235--3243, 2015.

\bibitem{MSG}
P.~Ochs and T.~Brox.
\newblock Object segmentation in video: a hierarchical variational approach for
  turning point trajectories into dense regions.
\newblock In {\em IEEE International Conference on Computer Vision}, pages
  1583--1590. IEEE, 2011.

\bibitem{ochs2012higher}
P.~Ochs and T.~Brox.
\newblock Higher order motion models and spectral clustering.
\newblock In {\em Computer Vision and Pattern Recognition (CVPR), 2012 IEEE
  Conference on}, pages 614--621. IEEE, 2012.

\bibitem{DAVIS2016}
F.~Perazzi, J.~Pont-Tuset, B.~McWilliams, L.~Van~Gool, M.~Gross, and
  A.~Sorkine-Hornung.
\newblock A benchmark dataset and evaluation methodology for video object
  segmentation.
\newblock In {\em Proceedings of the IEEE Conference on Computer Vision and
  Pattern Recognition}, pages 724--732, 2016.

\bibitem{perazzi2015fully}
F.~Perazzi, O.~Wang, M.~Gross, and A.~Sorkine-Hornung.
\newblock Fully connected object proposals for video segmentation.
\newblock In {\em Proceedings of the IEEE International Conference on Computer
  Vision}, pages 3227--3234, 2015.

\bibitem{objectness}
P.~O. Pinheiro, T.-Y. Lin, R.~Collobert, and P.~Doll{\'a}r.
\newblock Learning to refine object segments.
\newblock In {\em European Conference on Computer Vision}, pages 75--91.
  Springer, 2016.

\bibitem{DAVIS2017}
J.~Pont-Tuset, S.~Caelles, F.~Perazzi, A.~Montes, K.-K. Maninis, Y.~Chen, and
  L.~Van~Gool.
\newblock The 2018 davis challenge on video object segmentation.
\newblock {\em arXiv preprint arXiv:1803.00557}, 2018.

\bibitem{uNet}
O.~Ronneberger, P.~Fischer, and T.~Brox.
\newblock U-net: Convolutional networks for biomedical image segmentation.
\newblock In {\em International Conference on Medical image computing and
  computer-assisted intervention}, pages 234--241. Springer, 2015.

\bibitem{shankar2015video}
N.~Shankar~Nagaraja, F.~R. Schmidt, and T.~Brox.
\newblock Video segmentation with just a few strokes.
\newblock In {\em Proceedings of the IEEE ICCV}, pages 3235--3243, 2015.

\bibitem{sheikh2009background}
Y.~Sheikh, O.~Javed, and T.~Kanade.
\newblock Background subtraction for freely moving cameras.
\newblock In {\em 2009 IEEE 12th International Conference on Computer Vision},
  pages 1219--1225. IEEE, 2009.

\bibitem{shi1998motion}
J.~Shi and J.~Malik.
\newblock Motion segmentation and tracking using normalized cuts.
\newblock In {\em Computer Vision, 1998. Sixth International Conference on},
  pages 1154--1160. IEEE, 1998.

\bibitem{MotAdapt}
M.~Siam, C.~Jiang, S.~W. Lu, L.~Petrich, M.~Gamal, M.~Elhoseiny, and
  M.~J{\"{a}}gersand.
\newblock Video segmentation using teacher-student adaptation in a human robot
  interaction {(HRI)} setting.
\newblock {\em CoRR}, abs/1810.07733, 2018.

\bibitem{PDB}
H.~Song, W.~Wang, S.~Zhao, J.~Shen, and K.-M. Lam.
\newblock Pyramid dilated deeper convlstm for video salient object detection.
\newblock In {\em Proceedings of the European Conference on Computer Vision
  (ECCV)}, pages 715--731, 2018.

\bibitem{MpNet}
P.~Tokmakov, K.~Alahari, and C.~Schmid.
\newblock Learning motion patterns in videos.
\newblock In {\em 2017 IEEE Conference on Computer Vision and Pattern
  Recognition (CVPR)}, pages 531--539. IEEE, 2017.

\bibitem{visMem}
P.~Tokmakov, K.~Alahari, and C.~Schmid.
\newblock Learning video object segmentation with visual memory.
\newblock In {\em {IEEE} International Conference on Computer Vision, {ICCV}
  2017, Venice, Italy, October 22-29, 2017}, pages 4491--4500, 2017.

\bibitem{lsmo}
P.~Tokmakov, C.~Schmid, and K.~Alahari.
\newblock Learning to segment moving objects.
\newblock {\em International Journal of Computer Vision}, 127(3):282--301,
  2019.

\bibitem{torr1998geometric}
P.~H. Torr.
\newblock Geometric motion segmentation and model selection.
\newblock {\em Philosophical Transactions of the Royal Society of London A:
  Mathematical, Physical and Engineering Sciences}, 356(1740):1321--1340, 1998.

\bibitem{tron2007benchmark}
R.~Tron and R.~Vidal.
\newblock A benchmark for the comparison of 3-d motion segmentation algorithms.
\newblock In {\em 2007 IEEE conference on computer vision and pattern
  recognition}, pages 1--8. IEEE, 2007.

\bibitem{wang2018revisiting}
W.~Wang, J.~Shen, F.~Guo, M.-M. Cheng, and A.~Borji.
\newblock Revisiting video saliency: A large-scale benchmark and a new model.
\newblock In {\em Proceedings of the IEEE Conference on Computer Vision and
  Pattern Recognition}, pages 4894--4903, 2018.

\bibitem{wang2017super}
W.~Wang, J.~Shen, J.~Xie, and F.~Porikli.
\newblock Super-trajectory for video segmentation.
\newblock In {\em Proceedings of the IEEE International Conference on Computer
  Vision}, pages 1671--1679, 2017.

\bibitem{ags}
W.~Wang, H.~Song, S.~Zhao, J.~Shen, S.~Zhao, S.~C. Hoi, and H.~Ling.
\newblock Learning unsupervised video object segmentation through visual
  attention.
\newblock In {\em Proceedings of the IEEE Conference on Computer Vision and
  Pattern Recognition}, pages 3064--3074, 2019.

\bibitem{backpropagationThrougTime}
P.~J. Werbos.
\newblock Backpropagation through time: what it does and how to do it.
\newblock {\em Proceedings of the IEEE}, 78(10):1550--1560, 1990.

\bibitem{devildecoder}
Z.~Wojna, J.~R.~R. Uijlings, S.~Guadarrama, N.~Silberman, L.~Chen, A.~Fathi,
  and V.~Ferrari.
\newblock The devil is in the decoder.
\newblock In {\em British Machine Vision Conference 2017, {BMVC} 2017, London,
  UK, September 4-7, 2017}, 2017.

\bibitem{xiang2008video}
T.~Xiang and S.~Gong.
\newblock Video behavior profiling for anomaly detection.
\newblock {\em IEEE transactions on pattern analysis and machine intelligence},
  30(5):893--908, 2008.

\bibitem{MoNetECCV2018}
H.~Xiao, J.~Feng, G.~Lin, Y.~Liu, and M.~Zhang.
\newblock Monet: Deep motion exploitation for video object segmentation.
\newblock In {\em Proceedings of the IEEE Conference on Computer Vision and
  Pattern Recognition}, pages 1140--1148, 2018.

\bibitem{yilmaz2006object}
A.~Yilmaz, O.~Javed, and M.~Shah.
\newblock Object tracking: A survey.
\newblock {\em Acm computing surveys (CSUR)}, 38(4):13, 2006.

\bibitem{zhang2013video}
D.~Zhang, O.~Javed, and M.~Shah.
\newblock Video object segmentation through spatially accurate and temporally
  dense extraction of primary object regions.
\newblock In {\em Computer Vision and Pattern Recognition (CVPR), 2013 IEEE
  Conference on}, pages 628--635. IEEE, 2013.

\end{thebibliography}
}

\end{document}